\PassOptionsToPackage{hyphens}{url}
\documentclass{article}
\usepackage{spconf,amsmath,graphicx,hyperref}
\usepackage{amssymb}
\usepackage{booktabs}
\usepackage{multirow}
\usepackage{dsfont}
\usepackage[numbers,sort&compress]{natbib}
\hypersetup{hidelinks}

\renewcommand{\paragraph}[1]{\par\medskip\noindent\textbf{#1}\hspace{0.5em}\ignorespaces}

\title{BQDNA: Multi-Head Binary Spherical Quantization \\ for Genomic Sequence Modeling}
\name{
\textbf{Xudong Wang$^{2,*}$,
Siwei Cheng$^{1,*}$,
Haozhe Hu$^{2,3,\dagger}$
}}
\address{
$^{1}$University of Science and Technology of China\\
$^{2}$Institute of Digital Twin, Eastern Institute of Technology, Ningbo \\
$^{3}$Shanghai Jiao Tong University 
}

\begin{document}
\ninept
\maketitle

\begingroup
\renewcommand{\thefootnote}{*}
\footnotetext{Equal contribution. \qquad
$\dagger$ Corresponding author.}
\endgroup

\begin{abstract}
Decoding biological information from long, repetitive DNA remains challenging. Genomic language models (gLMs) often follow the conventional language-model pretrain-and-adapt paradigm. However, the lack of inherent semantic segmentation and prior rules makes hand-crafted tokenizers like k-mer and BPE suboptimal. We introduce context-aware \textbf{BQVAE} for scalar quantization of DNA and \textbf{BQDNA}, an autoregressive gLM supporting multi-head codebooks. Comprehensive experiments reveal that: (1) unique training dynamics of vector quantization in DNA space, including the importance of alignment between training corpus and downstream data. (2) superior performance of BQVAE over hand-crafted tokenizers and vanilla VQVAE. (3) remarkable cross-species generalization capability of BQDNA, which achieves state-of-the-art performance with merely 0.1B multi-species training for BQVAE and 1 epoch training on hg38 for BQDNA, outperforming gLMs pretrained with 10$\sim$100x more tokens. Our findings underscore the potential of learned vector quantization for genomic sequence representation. Code is available at \url{https://github.com/Sophie10001b/BQDNA}.
\end{abstract}



\section{Introduction}
Deoxyribonucleic acid (DNA), with 4 types of bases \{A, C, G, T\}, is the blueprint that encodes an organism's entire genetic information. Given DNA's fundamental role in biological language, numerous heuristic algorithms have been developed for DNA assembly, annotation, and higher-order association mining with diseases and drugs~\cite{perez-iratxeta2005G2D,culp2020Evolutionguideda}. Meanwhile, the continuous expansion of genomic databases and regulatory elements repositories (e.g., promoters, enhancers, and epigenetic markers)~\cite{meylan2020epd,meuleman2020Index,marakulina2023epifactors} has facilitated the widespread adoption of data-driven deep learning approaches for genomic and molecular analysis, which have already demonstrated their superiority over conventional heuristic algorithms across multiple downstream tasks~\cite{dealmeida2022DeepSTARR,jaganathan2019Predictinga,zhou2015Predicting,hu2023enhanced}. Large-scale self-supervised learning has further advanced gLM pretraining on unlabeled reference sequences for downstream genomic applications~\cite{wu2025GENERator}.

Amid the rapid development of Large Language Models (LLMs) and Genomic Language Models (gLMs), well-suited tokenization remains a critical bridge between the original char-based corpus and an LM's tokenized inputs~\cite{dagan2024Getting, tao2024Scaling}. Except for the simplest k-mer tokenization ($k\geqslant 1$, with $k=1$ as single nucleotide-based tokenization) that treats each adjacent k-length nucleotide chars as a single token~\cite{ji2021DNABERT,nguyen2023HyenaDNA,schiff2024Caduceus,dalla-torre2025NT,wu2025GENERator}, recent gLMs have incorporated with the byte-pair-encoding (BPE) to iteratively merge frequent subwords into larger tokens~\cite{zhou2023DNABERT2,fishman2025GENALM}, and further explored the learnable tokenization layer as a finer-grained and generalize preprocessing module~\cite{li2024VQDNA,qiao2024Model}. However, conventional k-mer and statistical-based tokenization face two fundamental limitations: (1) they cannot effectively incorporate domain-specific biological priors, and (2) as a simple preprocessing operation, it is difficult to guide them in capturing semantic patterns for downstream tasks. While recent VQ-based methods like VQDNA have introduced residual quantization (RQ)~\cite{lee2022Autoregressivea} to enhance sequence representations, the lack of specific investigations and details on the corpus for VQVAE training, sequence downsampling, and multi-head codebooks in gLM pretraining leaves the potential of VQ-based gLMs unclear.

\begin{figure}[t]
    \centering
    \includegraphics[trim=600 650 700 20, clip, width=0.9\linewidth]{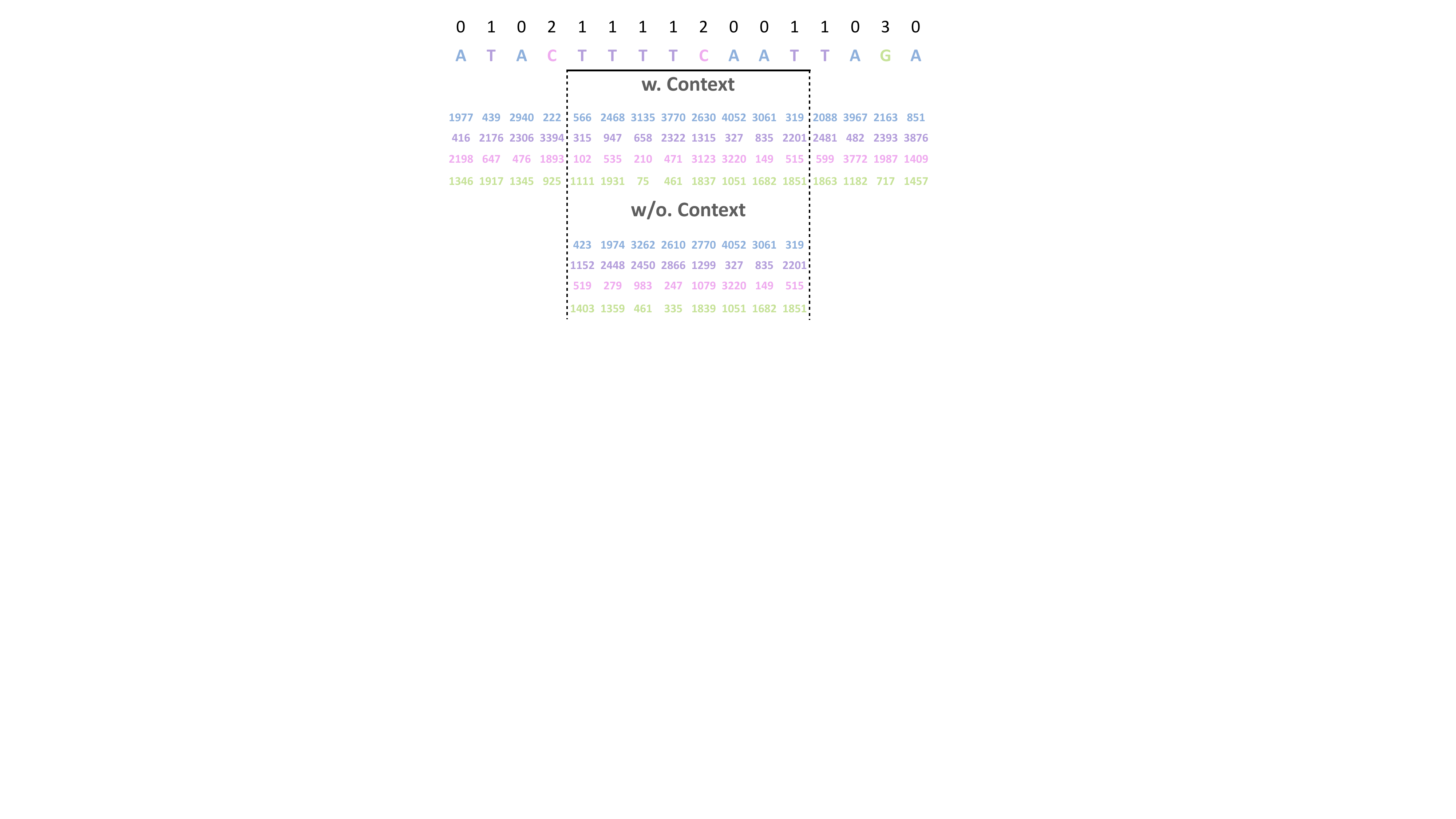}
    \caption{Example of four-head BQVAE tokenization with and without surrounding sequence context.}
    \label{tbl:tokenize_example}
    \vspace{-10pt}
\end{figure}

Building on the success of scalar quantization (SQ) in image discretization~\cite{mentzer2024finite}, we introduce a learnable DNA tokenization and pretraining framework with three contributions. (1) We develop \textbf{BQVAE} using multi-head binary spherical quantization (BSQ)~\cite{zhao2024Image} to expand and diversify the encoding space of nucleotide sequences. Figure~\ref{tbl:tokenize_example} illustrates that BQVAE can assign different multi-head codes to the same nucleotide segment depending on its surrounding sequence context. (2) We construct \textbf{BQDNA} as an autoregressive gLM that predicts multi-head codebook indices in parallel, enabling genomic pretraining on BQVAE representations. (3) We demonstrate the effectiveness and training efficiency of this framework through extensive experiments. BQVAE outperforms conventional hand-crafted tokenizers and vanilla VQVAE while requiring only 0.1B base pairs from a multi-species training corpus to transfer cross-species priors to a downstream gLM. Evaluations across more than 50 downstream datasets show that BQDNA matches or surpasses baselines while requiring only 2\% of the pretraining budget.
\section{Modeling Genomic with BQDNA}
Given a DNA sequence $\mathbf{S}$, BQVAE first maps it into latent features $\mathbf{Z}=\mathcal{E}(\mathbf{S})$ and independently quantizes the features across $H$ heads before reconstruction by $\mathcal{D}(\cdot)$. Instead of the nearest-neighbor codebook lookup used in conventional VQVAE ~\cite{vandenoord2017Neural}, we adopt binary spherical quantization (BSQ)~\cite{zhao2024Image}:
\begin{equation}
    \hat{\mathbf{z}}_i^h =
    \ell_2\text{-Norm}\left(
    \operatorname{sign}\left(
    \ell_2\text{-Norm}(\mathbf{z}_i^h)
    \right)\right).
\end{equation} 
The $d_e=256$ encoder features are projected to $D^\prime=48$ quantization dimensions, split into $H=4$ heads of $b=D^\prime/H=12$ bits. Each head defines $K=2^b=4096$ codes without explicit codebook lookup or optimization.

\begin{figure}[!h]
    \centering
    \includegraphics[trim=450 60 500 10, clip, width=1.0\linewidth]{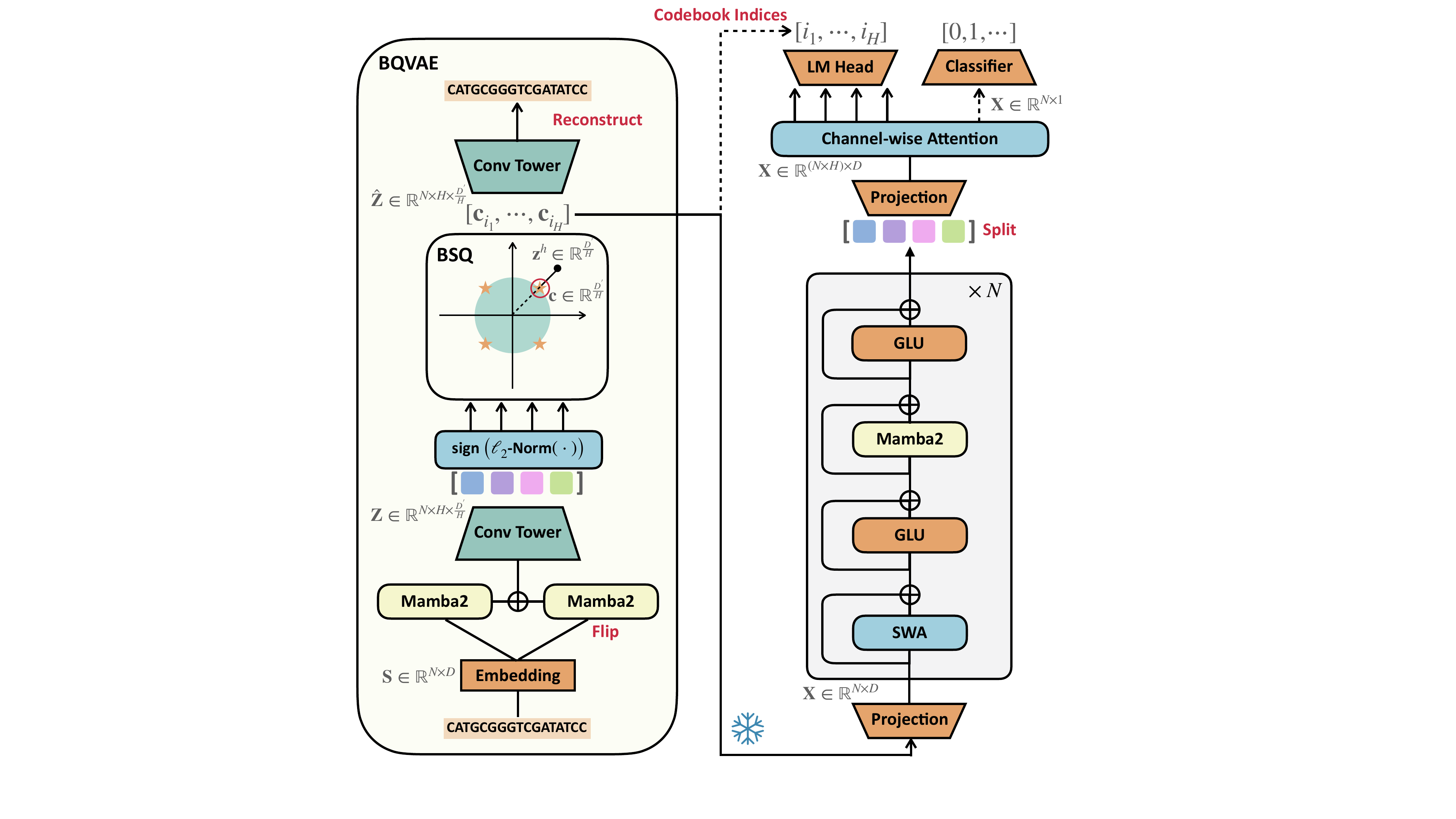}
    \caption{Overview of BQDNA. On the pretraining stage of BQDNA, BQVAE is frozen to transform the original sequences into codebook indices for causal language modeling.}
    \label{tbl:overview}
\end{figure}
Similar to VQDNA and VQVAE-based image generation approaches, BQDNA uses a two-stage pretraining paradigm. In the 1st stage, BQDNA trains a general BQVAE on functional element datasets that follow distributions analogous to downstream tasks, mapping raw sequences into a rich discrete latent space for subsequent genomic sequence modeling. In the subsequent phase, the frozen BQVAE acts as a standalone tokenizer to discretize the pretraining corpus into codebook indices, on which the language model (LM) directly performs causal language modeling (CLM). Figure~\ref{tbl:overview} illustrates BQDNA, and the following paragraphs elaborate on it.

\paragraph{BQ Tokenization for Sequences.}
Image tokenizers often use auxiliary training objectives~\cite{esser2021Taming}. A 24-bit RGB pixel has $2^{24}$ possible states, whereas a DNA base has $2^2$. With $H$ ordered heads and a shared codebook $\mathbf{C}$ of size $K$, each latent position has at most $K^H$ distinct code tuples. This multi-head design expands the code-combination space while retaining a compact prediction vocabulary of $K$ classes per head. To learn context-aware representations with balanced codebook utilization, we combine reconstruction with an entropy-based auxiliary loss, following LFQ and BSQ~\cite{yu2023Language,zhao2024Image}:
\begin{align*}
    &\mathcal{L}_{BQ} = \mathcal{L}_{rec} + \beta \mathcal{L}_{entropy}\\
    &\mathcal{L}_{entropy} = \mathbb{E}(H(\hat{\mathcal{Q}}(\mathbf{c}|\mathbf{z}))) - H(\mathbb{E}(\hat{\mathcal{Q}}(\mathbf{c}|\mathbf{z})))\\
    &\hat{\mathcal{Q}}(\mathbf{c}|\mathbf{z}) = \operatorname{softmax}(\mathbf{z}\mathbf{c}^\top / \tau)
\end{align*}
where $H(\cdot)$ denotes the entropy calculation, e.g.,
\begin{equation*}
    -\hat{\mathcal{Q}}(\mathbf{c}|\mathbf{z})\odot
    \log\hat{\mathcal{Q}}(\mathbf{c}|\mathbf{z}),
\end{equation*}
$\mathbb{E}(\cdot)$ denotes the expected value under the given distribution, which is similar to a reduced sum in this case, and $\beta$ and $\tau$ are hyperparameters for the entropy-based loss and distance calculation, respectively. The $\mathbb{E}(H(\hat{\mathcal{Q}}(\mathbf{c}|\mathbf{z})))$ is for the sample-wise entropy, and the second part in $\mathcal{L}_{entropy}$ is for batch-level diversity.

In our experiments, the reconstruction loss $\mathcal{L}_{rec}$ converges within the first 1k steps for both EMA-updated VQVAE with entropy regularization and the BSQ variant. Motivated by the distinction between reconstruction quality and downstream utility in image autoencoders~\cite{skorokhodov2025Improving}, we prioritize the functional content of the BQVAE training corpus to support subsequent genomic modeling. Specifically, we curate multi-species sequences of varying lengths from the Gener Tasks gene classification dataset~\cite{wu2025GENERator}, rather than using entire reference genomes containing abundant repetitive sequences. This selection focuses tokenizer training on sequence patterns relevant to downstream genomic annotation.

\paragraph{Predicting Multi-head Codebook Indices in $\mathcal{O}(N)$.}
Efficiently processing multi-head latent vectors at their original length is challenging during pretraining. To bound the complexity to $\mathcal{O}(N)$ for long-sequence modeling with base-level multi-head codebook retrieval, BQDNA employs a hybrid architecture that interleaves sliding window attention (SWA)~\cite{beltagy2020longformer} with 1k-bps window for causal modeling and 512-bps window for bidirectional downstream classification~\cite{dao2023FlashAttention2}, as well as linear attention modules~\cite{ren2024Samba,yang2024Parallelizing,wang2025Testtime}, with the latter directly implements the widely-adopted Mamba2 with data-dependency memory decay to ensure simplicity and efficiency~\cite{dao2024Transformers,wu2026data,hu2026beyond,hu2026wide,liu2026vica,Wu_2026_CVPR}. Furthermore, drawing from the form of multi-head VQVAE in image generation~\cite{lee2022Autoregressivea}, a channel-wise attention layer is set before the projection head to perform channel-level features interaction across all codebook heads $H$, which offers explicit guidance for predicting $H$ codebook indices $\hat{\mathbf{y}}$ from each fused token $\mathbf{x}$:
\begin{align*}
    &\mathbf{x}\in\mathbb{R}^{D} = [\mathbf{x}^1, \cdots, \mathbf{x}^H]\text{ , where}\ \mathbf{x}^H\in\mathbb{R}^{\frac{D}{H}}\\
    &\hat{\mathbf{y}} = [\hat{\mathbf{y}}^1, \cdots, \hat{\mathbf{y}}^H] = \operatorname{Attn.}([\mathbf{x}^1, \cdots, \mathbf{x}^H])
\end{align*}
Since the latent information in each head is independent in our multi-head BQVAE, there is no explicit positional encoding for the channel-level attention.

\begin{table*}[htbp]
    \centering
    \caption{Averaged MCC (in \%) performance on GUE and Nucleotide Transformer (revised) benchmarks across 3 runs; performance ranked in top-1 and top-2 is highlighted in \textbf{bold} and \underline{underline}, respectively. "--" denotes models that fail to converge on the dataset across 3 runs. Since DNABERT and NT-500M-1000G, marked by "*," fail to fine-tune on both GUE and NT benchmarks, the benchmark results from DNABERT-2's reports are introduced for comparison as baselines.}
    \aboverulesep=0ex
    \belowrulesep=0ex
    \resizebox{\linewidth}{!}{
        \begin{tabular}{l c|ccccccc|ccccc}
            \toprule
            \multirow{2}{*}{\textbf{Models}} & \multirow{2}{*}{\textbf{Train Tokens}} & \multicolumn{7}{c}{\textbf{GUE Benchmark}} & \multicolumn{5}{c}{\textbf{NT Benchmark}}\\
            \cmidrule(lr){3-9}\cmidrule(lr){10-14}
            & & EMP & TFM & TFH & PD & CPD & SSP & \textbf{Avg.} & EMP & ED & PD & SSP & \textbf{Avg.}\\
            \midrule
            DNABERT (6-mer)$^*$ & 122B & 49.10 & 56.43 & 64.17 & 81.70 & 71.81 & 84.07 & 60.69 & -- & -- & -- & -- & --\\
            DNABERT-2 & 262B & 56.04 & \textbf{73.68} & 63.38 & 81.77 & 67.02 & 88.01 & 65.93 & \textbf{56.79} & 51.66 & 79.21 & 86.29 & 64.87\\
            NT-500M-1000G$^*$ & 50B & 47.68 & 49.31 & 58.92 & 86.58 & 69.13 & 80.97 & 58.00 & -- & -- & -- & -- & --\\
            NTv2-100M & 300B & 51.79 & 65.68 & 61.23 & \underline{88.87} & \underline{73.19} & 87.84 & 63.95 & 55.12 & \underline{55.35} & \underline{81.32} & 96.75 & \underline{66.45}\\
            NTv2-500M & 300B & 57.63 & 71.59 & 67.36 & \textbf{91.39} & \textbf{74.74} & \underline{90.49} & 68.89 & \underline{56.74} & \textbf{56.53} & \textbf{84.01} & \textbf{97.06} & \textbf{67.98}\\
            GENA-LM & 275B & 47.01 & 64.69 & 64.38 & 81.12 & 67.16 & 76.55 & 60.63 & 56.01 & 51.78 & 76.58 & 80.86 & 63.11\\
            HyenaDNA & 5B & 47.69 & 65.40 & 64.28 & 78.53 & 68.76 & 77.23 & 60.90 & 52.70 & 51.24 & 79.41 & 74.36 & 60.60\\
            Caduceus-Ps & 52B & 55.76 & \underline{72.48} & 66.93 & 78.02 & 70.11 & 84.05 & 66.05 & 51.20 & 47.78 & 76.02 & 81.69 & 60.04\\
            \midrule
            BQDNA & & & & & & & & & & & & &\\
            \quad \emph{Hybrid} & \textbf{3.2B} & \textbf{68.93} & 72.44 & 67.56 & 84.32 & 70.20 & 89.79 & \textbf{71.96} & 54.93 & 49.08 & 80.17 & 96.76 & 65.46\\
            \quad \emph{Hybrid w/o. channel attn.} & 3.2B & \underline{66.24} & 71.33 & 67.76 & 83.67 & 70.06 & \textbf{90.93} & \underline{70.75} & 54.81 & 50.42 & 78.40 & \underline{96.79} & 65.26\\
            \quad \emph{Hybrid w. base} & 3.2B & 50.11 & 69.56 & \underline{67.89} & 81.27 & 70.75 & 86.93 & 64.13 & 50.36 & 49.60 & 79.08 & 92.90 & 62.15\\
            \quad \emph{Hybrid w. VQVAE} & 3.2B & 52.55 & 69.86 & 64.83 & 81.33 & 69.77 & 90.11 & 64.53 & 50.34 & 51.79 & 80.58 & 96.26 & 63.19 \\
            \quad \emph{Attn.} & 3.2B & 56.60 & 72.17 & \textbf{67.93} & 82.82 & 69.28 & 88.17 & 67.08 & 55.61 & 51.15 & 79.92 & 96.26 & 65.94\\
            \quad \emph{Attn. w. downsample} & 3.2B & 48.25 & 58.05 & 59.47 & 85.47 & 69.85 & 79.11 & 59.82 & 51.03 & 46.37 & 77.71 & 95.63 & 62.39\\
            \bottomrule
        \end{tabular}
    }
    \label{tbl:GUE_NT_benchmark}
\end{table*}

\paragraph{Training Details.}
Training follows two stages. We first train a 1.7M-parameter BQVAE on the Gene Task gene classification dataset for 3 epochs (approximately 9k steps), with encoder feature size $d_e=256$, quantization size $D^\prime=48$, codebook size $K=4096$, and $H=4$ heads. Its bidirectional encoder contains two Mamba2 blocks, while a symmetric 1D convolutional tower performs optional downsampling and upsampling. We then freeze BQVAE and pretrain a 110M-parameter BQDNA using causal language modeling. The backbone has hidden size $D=512$ and uses either 24 full-attention layers (\emph{Attn.}) or 12 SWA and 12 Mamba2 layers (\emph{Hybrid}). All variants are pretrained on human hg38 for 1 epoch (approximately 24k steps). Standard variants use $2^{10}$-bp inputs, whereas the 8$\times$ downsampling variant processes $2^{13}$ bp to produce $2^{10}$ latent tokens under a matched training budget.

\section{Experiments}

\paragraph{Baselines.}
We compare BQDNA with several pretrained gLMs of similar scale, including the recent 100M-parameter DNABERT-2~\cite{zhou2023DNABERT2}, Nucleotide Transformer v2-100M (NTv2)~\cite{dalla-torre2025NT}, and GENA-LM~\cite{fishman2025GENALM}. DNABERT~\cite{ji2021DNABERT} and Nucleotide Transformer-500M-1000G~\cite{dalla-torre2025NT}, both pretrained on human genomes, are included to compare the sample efficiency of BQ and vanilla k-mer tokenization. HyenaDNA~\cite{nguyen2023HyenaDNA}, Caduceus-Ps~\cite{schiff2024Caduceus}, and BQDNA without BQVAE serve as base-tokenization baselines. We further include the state-of-the-art NTv2-500M for a stronger comparison. Because no official VQDNA~\cite{li2024VQDNA} implementation is available, we include a BQDNA variant with a vanilla VQVAE tokenizer as an ablation.

\paragraph{Settings.}
All pretrained models are fine-tuned with AdamW~\cite{loshchilov2018decoupled} at a peak learning rate of 5e-5, with a 10\% linear warm-up over steps and no decay. We set the batch size to 32 for 5k update steps. To ensure stable performance, we evaluate each model every 200 steps and ensemble the top-5 checkpoints as the final result, repeating the process 3 times with different random seeds.

\paragraph{GUE Benchmark.}
We evaluate BQDNA on 27 of the 28 datasets in the Genome Understanding Evaluation (GUE)~\cite{zhou2023DNABERT2}, excluding Covid Variants Classification (CVC) because some methods exceeded our GPU memory. The 6 tasks comprise 1 Splice Site Prediction (SSP) dataset, 6 Promoter and Core Promoter Detection (PD \& CPD) datasets, 10 human and mouse Transcription Factor Prediction (TFH \& TFM) datasets, and 10 yeast Epigenetic Marker Prediction (EMP) datasets. We report Matthews Correlation Coefficient (MCC). The selected datasets span human, mouse, and yeast sequences, enabling assessment of cross-species generalization.

As shown in Table~\ref{tbl:GUE_NT_benchmark}, the hybrid BQDNA model achieves the highest average MCC on GUE among the evaluated models. BQDNA uses the minimal seen tokens during pretraining. With the same hybrid backbone, BQVAE raises the average MCC from 64.13 with base tokenization and 64.53 with vanilla VQVAE to 71.96, highlighting the benefit of its learned token representations. It also exceeds NTv2-500M, the strongest external baseline on GUE, which obtains an average MCC of 68.89. Channel-wise attention further improves the average MCC from 70.75 to 71.96. The advantage is especially pronounced on EMP, where BQVAE reaches 68.93 MCC versus 50.11 for base tokenization and 52.55 for VQVAE under the same hybrid backbone. BQDNA's robust cross-species generalization on EMP further highlights its strong ability to learn efficient multi-species sequence quantization, suggesting that a sufficiently trained BQVAE tokenizer can enable a single-species pre-trained gLM to generalize well to downstream multi-species tasks. The 8$\times$ downsampling variant also surpasses NT-500M-1000G with 6-mer tokenization on GUE, particularly in EMP and TFM. These results demonstrate BQVAE's advantages over hand-crafted segmentation.

\paragraph{NT (revised) Benchmark.}
The second Benchmark introduced by Nucleotide Transformer~\cite{dalla-torre2025NT} is another touchstone for evaluating gLMs, focusing mainly on functional and regulatory elements within the human genome. NT benchmark consists of 4 tasks with 18 datasets: 10 datasets in Epigenetic Marker Prediction, 2 datasets in Enhancer Detection (ED), 3 datasets in Promoter Detection, and 3 datasets in Splice Site Prediction, with MCC as the evaluation metric. We extract all sequences on chromosome 10 (chr10) from training sets to construct the held-out validation set. As shown on the right side of Table~\ref{tbl:GUE_NT_benchmark}, both BQDNA variants maintain competitive performance on NT with the fewest pretraining tokens. The gains from BQVAE extend to this Benchmark: with the same hybrid backbone, the average MCC improves from 62.15 with base tokenization and 63.19 with vanilla VQVAE to 65.46. These consistent gains across both benchmarks support BQVAE as an effective tokenizer beyond a single evaluation suite. The 8$\times$ downsampling variant exhibits a smaller MCC reduction on NT than on GUE relative to the full-resolution attention variant, indicating task-dependent sensitivity to tokenization resolution.

\paragraph{Frozen-Embedding Regulatory Classification.}
To assess representation discriminability, we train a supervised XGBoost~\cite{chen2016xgboost} classifier on mean-pooled embeddings from each frozen gLM using downstream labels~\cite{feng2024Benchmarking}. We evaluate all models on 3 tasks with 10 datasets: 2 datasets for detecting human methylation modification (i.e., EMP tasks), 4 datasets for Human Promoter Detection (PDH) across different cell lines, and 4 datasets for Multi-species Promoter Detection (PDM), using MCC as the evaluation metric. As shown in Table~\ref{tbl:zero_shot_benchmark}, hybrid BQDNA achieves the highest mean MCC on EMP and multi-species PDM, as well as the highest overall mean across the ten evaluated datasets among models with complete results. Its 63.74 MCC on multi-species PDM exceeds the strongest listed baseline, Caduceus-Ps, at 60.77; its overall mean is 66.94 compared with 66.68 for that baseline. These results highlight BQVAE's ability to provide cross-species representations for a gLM backbone pretrained on hg38 alone. These frozen-embedding results complement the fine-tuning benchmarks, indicating that BQDNA's representations support supervised classification without adapting the gLM itself to downstream labels.
\begin{table}[!t]
    \centering
    \caption{Averaged MCC (in \%) performance on regulatory classification tasks using frozen gLM embeddings across 3 runs; each pretrained model's final hidden states are fed into an XGBoost model to obtain the prediction results. "--" denotes models that fail to process sequences with different lengths and 2k-bps contexts.}
    \begin{tabular}{l cccc}
        \toprule
        \textbf{Models} & EMP & PDH & PDM & \textbf{Avg.}\\
        \midrule
        DNABERT-2 & 26.09 & 83.95 & 51.44 & 59.38\\
        NTv2-100M & 35.50 & 83.61 & 55.45 & 62.72\\
        NTv2-500M & 37.17 & \underline{84.67} & 56.77 & 64.01\\
        GENA-LM & \underline{40.57} & -- & 59.14 & --\\
        HyenaDNA & 30.36 & 84.05 & 57.78 & 62.80\\
        Caduceus-Ps & 39.31 & \textbf{86.28} & 60.77 & \underline{66.68}\\
        \midrule
        BQDNA & & & &\\
        \quad \emph{Hybrid} & \textbf{41.47} & 82.86 & \textbf{63.74} & \textbf{66.94}\\
        \quad \emph{Attn.} & 40.06 & 82.74 & \underline{63.30} & 66.43\\
        \bottomrule
    \end{tabular}
    \label{tbl:zero_shot_benchmark}
\end{table}

\paragraph{Ablation Studies.}
Because the downstream backbones are trained from scratch, this comparison examines how pretrained tokenizers shape subsequent learning on downstream tasks without relying on knowledge from a pretrained language-model backbone.

\begin{figure}[!t]
    \centering
    \includegraphics[width=\linewidth]{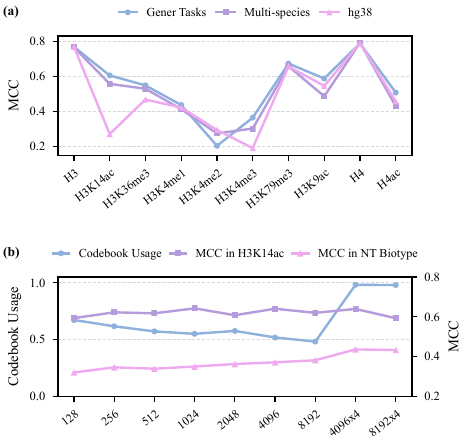}
    \caption{Combined ablation studies of BQVAE. (a) MCC on GUE EMP tasks for BQVAEs trained with different corpora. All \emph{Hybrid} models are trained from scratch, while the BQVAE variants using human-only or multi-species reference genomes are trained for 10k steps. (b) Effects of codebook size and head count ($K\times H$) on H3K14ac MCC, codebook usage, and NT Biotype MCC measured with supervised XGBoost on frozen embeddings.}
    \label{fig:combined_ablation}
\end{figure}

Benchmarks on various datasets show that context-aware BQVAE can transfer rich cross-species knowledge to the gLM backbone using only a limited multi-species training corpus, compared with naive base tokenization. To systematically examine how the training corpus affects BQVAE's cross-species transferability, we introduce 3 BQDNA variants with BQVAE trained on different datasets (Gener Tasks, multi-species, and human-only reference genomes) and train them from scratch on the GUE EMP task. As the validation curves illustrated in Figure~\ref{fig:combined_ablation}(a), BQVAE with a multi-species corpus can provide substantially more informative prior knowledge through its tokenization patterns than its hg38-only counterpart, yielding significant performance gains on cross-species tasks. This finding not only confirms BQVAE's sensitivity to the training data distribution, but also suggests that while BQVAE requires a small amount of training data in DNA space for lossless reconstruction, careful corpus curation can markedly improve downstream generalizability.

Beyond the ablation on training corpus, codebook size represents another critical hyperparameter in BQVAE. To evaluate this trade-off, we report test-set performance on H3K14ac trained from scratch, step-wise averaged codebook utilization during H3K14ac training, and supervised multi-class classification performance using frozen embeddings across 4 distinct functional elements in NT Benchmark (i.e., the NT Biotype dataset) for BQVAE with different codebook sizes. Figure~\ref{fig:combined_ablation}(b) shows that increasing codebook size alone does not consistently improve downstream MCC, motivating a balanced choice of codebook size and head count. The multi-head configurations achieve higher codebook utilization and stronger functional-element discriminability, supporting multiple heads to enrich token representations. BQDNA therefore adopts $K=4096$ codes per head with $H=4$ heads.

\paragraph{Analysis of Tokenization Behavior.}
To analyze BQVAE tokenization behavior, we examine the Gene Task gene classification dataset used for BQVAE training. Specifically, we compute the mean utilization frequency of each codebook token across all samples. As a baseline, we use the BPE tokenizer in DNABERT-2, a hand-crafted, frequency-based approach. The comparison shows that BPE exhibits a significant long-tail distribution in token utilization, whereas BQVAE achieves balanced token activation with only minimal low-probability outliers. Such allocation enables more comprehensive token representation learning in the subsequent gLM backbone, highlighting BQVAE's advantages in both token usage and further codebook extension.

\section{Conclusion}
In this paper, we investigate the potential of multi-head BSQ for context-aware DNA tokenization and propose BQDNA, a bidirectional BQVAE quantizer based on Mamba2 and a symmetric convolutional tower, with a gLM framework that supports autoregressive prediction over multi-head codebooks. Extensive experiments over 50+ datasets reveal that BQVAE can provide valuable cross-species priors for BQDNA's backbone with a minimal proper training corpus, which pushes BQDNA to achieve remarkable downstream performance with merely one epoch of pretraining on hg38, only 2\% of the training tokens compared with baselines. These benchmarking results further highlight BQVAE's effectiveness as a learnable DNA tokenization, as well as its notable sample efficiency compared with hand-crafted strategies and vanilla VQVAE.



\makeatletter
\let\BQDNAorigthebibliography\thebibliography
\renewcommand{\thebibliography}[1]{%
  \BQDNAorigthebibliography{#1}%
  \setlength{\itemsep}{0pt}%
  \setlength{\parsep}{0pt}%
}
\makeatother

\renewcommand{\bibsection}{\section{References}}
\clearpage
\section{Compliance with Ethical Standards}
This study used publicly available genomic reference sequences and benchmark datasets. No new human or animal samples were collected. No additional ethical approval was required for this study.


\bibliographystyle{IEEEbibNatCompact}
\bibliography{main}

\end{document}